\documentclass[11pt,a4paper]{article}

\usepackage[T1]{fontenc}
\usepackage[utf8]{inputenc}
\usepackage{mathptmx}
\usepackage[margin=2.5cm]{geometry}
\usepackage{amsmath,amssymb,amsthm}
\usepackage{physics}
\usepackage{bm}
\usepackage{graphicx}
\usepackage{booktabs}
\usepackage{multirow}
\usepackage{array}
\usepackage{xcolor}
\usepackage{hyperref}
\usepackage{authblk}
\usepackage{natbib}
\usepackage{tikz}
\usetikzlibrary{arrows.meta,decorations.pathreplacing,calc}
\usepackage[ruled,vlined,linesnumbered]{algorithm2e}
\usepackage{microtype}
\usepackage{caption}
\usepackage{subcaption}
\usepackage{enumitem}
\usepackage{float}

\graphicspath{{figs/}}

\definecolor{denseblue}{RGB}{26,58,92}
\definecolor{mpored}{RGB}{176,48,96}
\definecolor{mpoorange}{RGB}{212,84,26}
\definecolor{mpopurple}{RGB}{108,52,131}
\definecolor{mpogreen}{RGB}{26,122,74}
\definecolor{linkblue}{RGB}{30,80,160}

\hypersetup{colorlinks=true,linkcolor=linkblue,citecolor=linkblue,urlcolor=linkblue}

\newcommand{\mpomat}[1]{\mathbf{#1}}
\newcommand{\R}{\mathbb{R}}

\newcommand{\calA}{\mathcal{A}}
\newcommand{\diag}{\operatorname{diag}}

\DeclareMathOperator*{\argmax}{arg\,max}

\theoremstyle{definition}
\newtheorem{definition}{Definition}

\title{\textbf{Compressing Transformer Language Models\\
		via Matrix Product Operator Decomposition}\\[0.5em]
	{\large A Case Study on PicoGPT}}

\author[1]{Younes Javanmard}
\affil[1]{Institut f\"{u}r Theoretische Physik, Leibniz Universit\"{a}t Hannover,
	Appelstra\ss{}e~2, 30167 Hannover, Germany\\
	\small\texttt{younes.javanmard@itp.uni-hannover.de}}

\author[2]{Tanmoy Pandit}
\affil[2]{VTT Technical Research Centre of Finland,
	Tietotie~3, Espoo, Finland\\
}

\author[3]{Masoud Mardani}
\affil[3]{National High Magnetic Field Laboratory, Tallahassee, FL 32310, USA\\
}

\date{\today}

\begin{document}
	
	\maketitle
	
	\begin{abstract}
		\noindent\textbf{Background.}
		Transformer-based language models achieve state-of-the-art performance across
		natural language tasks, but their quadratic parameter scaling with hidden
		dimension makes deployment on resource-constrained hardware expensive.
		Existing compression methods --- including pruning, quantization, and
		low-rank factorization --- treat all weight structures uniformly and offer
		limited control over the approximation error. Matrix Product
		Operator (MPO) decomposition, a tensor network technique originally developed
		for quantum many-body simulations, provides a principled alternative:
		it factorizes weight matrices into chains of low-rank cores whose approximation
		quality is governed by a single interpretable hyperparameter, the bond
		dimension~$\chi$.
		
		\noindent\textbf{Methods.}
		We replace every \texttt{nn.Linear} layer in PicoGPT --- a GPT-2-style
		character-level language model with ${\sim}1$\,M parameters --- with an
		\texttt{MPOLinear} module whose weight is parameterised as an MPO chain.
		Cores are initialized either via the TT-SVD algorithm applied to pretrained
		dense weights or from random initialisations. All cores are stored
		as standard \texttt{nn.Parameter} tensors; gradient flow through the
		\texttt{tensordot} contraction chain is handled automatically by PyTorch
		autograd, requiring no custom backward pass. We derive balanced factorization
		schemes for each of the five distinct weight shapes in PicoGPT and evaluate
		bond dimensions $\chi \in \{4, 8, 16, 32\}$ on the Tiny Shakespeare corpus.
		
		\noindent\textbf{Results.}
		MPO compression achieves up to $13\times$ parameter compression per transformer
		block at $\chi=4$. At $\chi=16$ --- with 191\,872 parameters versus the dense
		baseline of 1\,020\,224 --- the model retains $97.7\%$ of the baseline token
		accuracy ($51.6\%$ vs.\ $52.8\%$), a gap of $1.2$ percentage points.
		Per-layer reconstruction error decreases systematically with increasing bond
		dimension and is consistently lower for three-site ($L=3$) factorisations than
		for two-site ($L=2$) ones at the same $\chi$. Under the parameter-efficiency
		proxy used in this work, the MPO model at $\chi=8$ attains the highest score.
		
		\noindent\textbf{Conclusions.}
		MPO compression provides a theoretically grounded and practically accessible
		route to transformer parameter compression with explicit control over the
		accuracy--compression trade-off. Our open-source PyTorch implementation
		is fully compatible with standard training pipelines and requires no
		modifications to the model training loop. Our current implementation demonstrates parameter compression and training compatibility;
		realising inference-time memory and FLOP savings will require direct structured contractions
		that avoid reconstructing the dense weight. These results suggest that MPO
		parameterization is a promising alternative to standard low-rank and
		tensor-factorized compression approaches, particularly when interpretable
		compression control and quantum-physics-motivated analysis are desired.
		
	\end{abstract}
	
	\tableofcontents
	\vspace{1em}
	\hrule
	\vspace{1em}
	
	\section{Introduction}
	
	Transformer language models have achieved state-of-the-art performance across a broad range of natural language tasks~\citep{vaswani2017attention,radford2019language,brown2020language}. However, their parameter counts scale quadratically with the hidden dimension, making deployment on resource-constrained hardware expensive. A key empirical observation motivating compression is that neural network weights are heavily redundant~\citep{denil2013predicting}, suggesting that low-dimensional structure is present. A variety of compression techniques have been proposed, including weight pruning~\citep{han2015deep}, CP-decomposition of convolutional filters~\citep{lebedev2015speeding}, knowledge distillation~\citep{hinton2015distilling}, and low-rank factorisation~\citep{hu2022lora}.
	
	An alternative perspective, imported from quantum information theory, represents weight tensors as \emph{Matrix Product Operators} (MPOs)~\citep{oseledets2011tensor,kolda2009tensor,novikov2015tensorizing}. An MPO factorises a high-dimensional weight tensor into a chain of low-rank \emph{cores}, each carrying only $\mathcal{O}(\chi^2 d^2)$ parameters when the local physical dimensions are of order $d$ and the bond dimension is $\chi$. The bond dimension is the single knob that controls the accuracy--compression trade-off: increasing $\chi$ systematically recovers the dense weight.
	
	This framework is familiar in the quantum many-body community under the name Tensor Train (TT)~\citep{oseledets2011tensor} or Matrix Product State (MPS) when applied to vectors; the operator variant (MPO) is the natural choice for weight matrices. MPO-based neural compression was pioneered by~\citet{novikov2015tensorizing} for fully-connected layers and subsequently extended to recurrent architectures~\citep{yang2017tensor,tjandra2017compressing}, transformers~\citep{gao2020compressing}, and more recently to large language models by~\citet{tomut2024compactifai}, who reported substantial compression of LLaMA-2~7B with a relatively modest accuracy drop.
	
	In this paper we apply MPO compression to \textbf{PicoGPT}~\citep{osborne2026picogpt}, a pedagogical GPT-2-style implementation written from scratch in Julia following the transformer architecture of~\citet{vaswani2017attention}, which we re-implement in PyTorch to enable gradient-based fine-tuning of the compressed cores. The MPO cores are treated as first-class \texttt{nn.Parameter} tensors with gradients flowing through the contraction chain automatically --- an approach related to the MPO arithmetic framework developed for quantum circuit compilation~\citep{javanmard2026cdmpo}. Our contributions are:
	\begin{enumerate}[itemsep=2pt]
		\item A clean, fully autograd-compatible MPO layer (\texttt{MPOLinear}) that replaces any \texttt{nn.Linear} without custom backward code.
		\item Factorisation schemes for all five weight shapes in PicoGPT, derived from balanced MPO design principles.
		\item A systematic benchmark comparing dense and MPO-parameterised models across bond dimensions $\chi \in \{4,8,16,32\}$ on character-level Shakespeare prediction.
		\item An analysis of reconstruction error, training dynamics, and the accuracy--compression Pareto frontier.
	\end{enumerate}
	
	\section{Background}
	
	\subsection{Tensor Train and Matrix Product Operators}
	
	\begin{definition}[Tensor Train]
		A tensor $\mathcal{T} \in \R^{n_1 \times n_2 \times \cdots \times n_L}$ is said to be in \emph{Tensor Train} (TT) format with bond dimensions $(\chi_1, \ldots, \chi_{L-1})$ if it can be written as
		\begin{equation}
			\mathcal{T}_{i_1 i_2 \cdots i_L}
			= \sum_{\alpha_1, \ldots, \alpha_{L-1}}
			\calA^{(1)}_{1, i_1, \alpha_1}\,
			\calA^{(2)}_{\alpha_1, i_2, \alpha_2}\,
			\cdots\,
			\calA^{(L)}_{\alpha_{L-1}, i_L, 1},
			\label{eq:tt}
		\end{equation}
		where $\calA^{(l)} \in \R^{\chi_{l-1} \times n_l \times \chi_l}$ are the \emph{TT-cores}, with boundary conditions $\chi_0 = \chi_L = 1$.
	\end{definition}
	
	\begin{definition}[Matrix Product Operator]
		A weight matrix $\mpomat{W} \in \R^{\mathrm{out} \times \mathrm{in}}$ is a \emph{Matrix Product Operator} (MPO) if, after reshaping $\mathrm{out} = \prod_{l=1}^L d_l^\mathrm{out}$ and $\mathrm{in} = \prod_{l=1}^L d_l^\mathrm{in}$, it can be expressed as
		\begin{equation}
			W_{(i_1\cdots i_L),(j_1\cdots j_L)}
			=
			\sum_{\alpha_1,\ldots,\alpha_{L-1}}
			\calA^{(1)}_{1,\, i_1,\, j_1,\, \alpha_1}\,
			\calA^{(2)}_{\alpha_1,\, i_2,\, j_2,\, \alpha_2}\,
			\cdots\,
			\calA^{(L)}_{\alpha_{L-1},\, i_L,\, j_L,\, 1},
			\label{eq:mpo}
		\end{equation}
		where $\calA^{(l)} \in \R^{\chi_{l-1} \times d_l^\mathrm{out} \times d_l^\mathrm{in} \times \chi_l}$.
	\end{definition}
	
	For a physics-oriented introduction to MPS and MPO see~\citet{orus2014practical}. Tensor networks were first applied to supervised machine learning by~\citet{stoudenmire2016supervised}, who showed that MPS can learn relevant features from image data.
	
	The exact number of parameters in an $L$-site MPO is
	\begin{equation}
		N_\text{MPO}
		=
		\chi_1 d_1^\mathrm{out} d_1^\mathrm{in}
		+ \sum_{l=2}^{L-1} \chi_{l-1}\chi_l\, d_l^\mathrm{out} d_l^\mathrm{in}
		+ \chi_{L-1} d_L^\mathrm{out} d_L^\mathrm{in},
		\label{eq:params_general}
	\end{equation}
	with $N_\text{dense} = \mathrm{out}\times\mathrm{in}$ for the full dense matrix. In the common special case of a uniform bond dimension $\chi_1=\cdots=\chi_{L-1}=\chi$, this becomes
	\begin{equation}
		N_\text{MPO}
		=
		\chi\, d_1^\mathrm{out} d_1^\mathrm{in}
		+ \sum_{l=2}^{L-1} \chi^2 d_l^\mathrm{out} d_l^\mathrm{in}
		+ \chi\, d_L^\mathrm{out} d_L^\mathrm{in}.
		\label{eq:params}
	\end{equation}
	For balanced factorizations with bounded local dimensions and fixed bond dimension, the MPO parameter count grows only linearly in the number of sites $L$, whereas the dense parameter count grows multiplicatively with the full input and output dimensions.
	
	\subsection{TT-SVD Algorithm}
	
	The TT-SVD algorithm~\citep{oseledets2011tensor} converts a dense weight matrix into an MPO via a sequential SVD sweep:
	
	\begin{algorithm}[H]
		\DontPrintSemicolon
		\caption{TT-SVD for MPO compression}
		\KwIn{$\mpomat{W} \in \R^{\mathrm{out} \times \mathrm{in}}$, factorisations $(d_l^\mathrm{out})$, $(d_l^\mathrm{in})$, maximal bond $\chi$}
		\KwOut{MPO cores $\{\calA^{(l)}\}_{l=1}^{L}$}
		
		Reshape $\mpomat{W}$ into an interleaved tensor $\mathcal{T}$ of shape $(d_1^\mathrm{out}, d_1^\mathrm{in}, \ldots, d_L^\mathrm{out}, d_L^\mathrm{in})$\;
		$\text{bond\_left} \leftarrow 1$\;
		$\mathcal{R} \leftarrow \mathcal{T}$ \tcp*{running remainder}
		\For{$l = 1, \ldots, L-1$}{
			Unfold: $\mpomat{M} \leftarrow \text{reshape}(\mathcal{R},\; \text{bond\_left} \cdot d_l^\mathrm{out} \cdot d_l^\mathrm{in},\; -1)$\;
			Compute truncated SVD: $\mpomat{M} \approx \mpomat{U}\,\diag(\bm{s})\,\mpomat{V}^\top$,\; $r \leftarrow \min(\chi, \mathrm{rank}(\mpomat{M}))$\;
			$\calA^{(l)} \leftarrow \text{reshape}(\mpomat{U}_{:,\,1:r},\; \text{bond\_left},\, d_l^\mathrm{out},\, d_l^\mathrm{in},\, r)$\;
			$\mathcal{R} \leftarrow \diag(\bm{s}_{1:r})\,\mpomat{V}_{:,\,1:r}^\top$,\; $\text{bond\_left} \leftarrow r$\;
		}
		$\calA^{(L)} \leftarrow \text{reshape}(\mathcal{R},\; \text{bond\_left},\, d_L^\mathrm{out},\, d_L^\mathrm{in},\, 1)$\;
	\end{algorithm}
	
	Let $\sigma_{l,r}$ denote the singular values discarded at unfolding step $l$. Then the TT-SVD approximation satisfies the standard Frobenius-norm bound~\citep{oseledets2011tensor}
	\begin{equation}
		\|\mpomat{W} - \widehat{\mpomat{W}}\|_F^2
		\le
		\sum_{l=1}^{L-1} \sum_{r>\chi_l} \sigma_{l,r}^2.
		\label{eq:error_bound}
	\end{equation}
	In particular, the reconstruction error decreases monotonically as the admissible bond dimensions increase.
	
	\subsection{Gradient Flow Through MPO Layers}
	
	During training, the weight matrix is reconstructed on each forward pass via the contraction
	\begin{equation}
		\widehat{\mpomat{W}} = \operatorname{contract}\!\left(\calA^{(1)},\calA^{(2)},\ldots,\calA^{(L)}\right).
		\label{eq:contract}
	\end{equation}
	Writing the reconstructed weight explicitly as
	\begin{equation}
		\widehat{\mpomat{W}}_{i_1\cdots i_L,\; j_1\cdots j_L}
		=
		\sum_{\alpha_1,\ldots,\alpha_{L-1}}
		\calA^{(1)}_{1,i_1,j_1,\alpha_1}
		\calA^{(2)}_{\alpha_1,i_2,j_2,\alpha_2}
		\cdots
		\calA^{(L)}_{\alpha_{L-1},i_L,j_L,1},
	\end{equation}
	the gradient of the loss $\mathcal{L}$ with respect to the $l$-th core is obtained by contracting the loss gradient with the left and right environments:
	\begin{equation}
		\frac{\partial \mathcal{L}}{\partial \calA^{(l)}_{\alpha_{l-1} i_l j_l \alpha_l}}
		=
		\sum_{\substack{
				i_1,\dots,i_{l-1},\,i_{l+1},\dots,i_L\\
				j_1,\dots,j_{l-1},\,j_{l+1},\dots,j_L\\
				\alpha_1,\dots,\alpha_{l-2},\,\alpha_{l+1},\dots,\alpha_{L-1}
		}}
		L^{(l)}_{\alpha_{l-1};\, i_1\cdots i_{l-1},\, j_1\cdots j_{l-1}}
		\;
		\frac{\partial \mathcal{L}}{\partial \widehat{\mpomat{W}}_{i_1\cdots i_L,\; j_1\cdots j_L}}
		\;
		R^{(l)}_{\alpha_l;\, i_{l+1}\cdots i_L,\, j_{l+1}\cdots j_L},
		\label{eq:gradient}
	\end{equation}
	where
	\begin{equation}
		L^{(l)}_{\alpha_{l-1};\, i_1\cdots i_{l-1},\, j_1\cdots j_{l-1}}
		=
		\sum_{\alpha_1,\dots,\alpha_{l-2}}
		\calA^{(1)}_{1,i_1,j_1,\alpha_1}
		\calA^{(2)}_{\alpha_1,i_2,j_2,\alpha_2}
		\cdots
		\calA^{(l-1)}_{\alpha_{l-2},i_{l-1},j_{l-1},\alpha_{l-1}},
	\end{equation}
	and
	\begin{equation}
		R^{(l)}_{\alpha_l;\, i_{l+1}\cdots i_L,\, j_{l+1}\cdots j_L}
		=
		\sum_{\alpha_{l+1},\dots,\alpha_{L-1}}
		\calA^{(l+1)}_{\alpha_l,i_{l+1},j_{l+1},\alpha_{l+1}}
		\cdots
		\calA^{(L)}_{\alpha_{L-1},i_L,j_L,1}.
	\end{equation}

	Thus, the gradient has the standard left-environment / local-core / right-environment contraction structure familiar from single-site tensor-network optimization methods such as DMRG~\citep{white1992density}. In PyTorch, this is computed automatically by autograd through the \texttt{tensordot}$\to$\texttt{permute}$\to$\texttt{reshape} chain in~\eqref{eq:contract}; no custom backward pass is needed.

	\section{Architecture and Implementation}
	
	\subsection{PicoGPT Architecture}
	
	PicoGPT~\citep{osborne2024picogpt} is a character-level GPT-2-style language model implemented in pure Julia without automatic differentiation. Table~\ref{tab:arch} lists its default hyperparameters. The architecture follows~\citet{radford2019language}: pre-norm residual blocks, sinusoidal positional encoding (no learnable PE), scaled dot-product multi-head causal self-attention, and a two-layer FFN with ReLU activation.
	
	\begin{table}[H]
		\centering
		\caption{PicoGPT default hyperparameters.}
		\label{tab:arch}
		\begin{tabular}{llr}
			\toprule
			Parameter & Description & Value \\
			\midrule
			$V$ & Vocabulary size (character-level) & 65 \\
			$D$ & Embedding dimension ($d_\text{model}$) & 128 \\
			$H$ & Attention heads & 4 \\
			$N$ & Transformer layers & 4 \\
			$T$ & Context length & 256 \\
			$d_\text{ff}$ & FFN hidden dimension ($4D$) & 512 \\
			\midrule
			Total & Dense parameters & ${\sim}$1\,020\,000 \\
			\bottomrule
		\end{tabular}
	\end{table}
	
	Each transformer block contains six linear layers:
	$\mpomat{W}_Q, \mpomat{W}_K, \mpomat{W}_V \in \R^{D \times D}$ (query/key/value projections),
	$\mpomat{W}_O \in \R^{D \times D}$ (attention output),
	$\mpomat{W}_1 \in \R^{d_\text{ff} \times D}$ (FFN up-projection), and
	$\mpomat{W}_2 \in \R^{D \times d_\text{ff}}$ (FFN down-projection).
	Additionally, the language-model head $\mpomat{W}_\text{LM} \in \R^{V \times D}$ maps hidden states to logits.
	
	\paragraph{Scope of compression.}
	In the present implementation, MPO parameterisation is applied only to affine weight
	matrices associated with linear projections, namely the attention projections
	$\mpomat{W}_Q,\mpomat{W}_K,\mpomat{W}_V,\mpomat{W}_O$, the feed-forward layers
	$\mpomat{W}_1,\mpomat{W}_2$, and the language-model head $\mpomat{W}_{\mathrm{LM}}$.
	Embedding tables, layer-normalization parameters, biases, and the sinusoidal positional
	encoding remain in their standard dense form. Consequently, the reported parameter counts
	reflect partial model compression rather than a fully tensorised architecture.
	
	\subsection{MPO Factorisation Schemes}
	
	For each linear layer shape, we choose a factorization $(d_l^\text{out}), (d_l^\text{in})$ that (a) produces reasonably balanced physical dimensions across sites and (b) yields an integer factorization of the output and input sizes. Table~\ref{tab:factorisations} summarizes our choices.
	
	\begin{table}[H]
		\centering
		\caption{MPO factorisation schemes for each PicoGPT linear layer. $\chi$ denotes the bond dimension.}
		\label{tab:factorisations}
		\begin{tabular}{l r r c l l r r}
			\toprule
			Layer & out & in & $L$ & $(d_l^\text{out})$ & $(d_l^\text{in})$ & Dense & MPO ($\chi=8$) \\
			\midrule
			$\mpomat{W}_{Q/K/V/O}$ & 128 & 128 & 2 & $[8,16]$   & $[8,16]$   & 16\,384 & 2\,560 \\
			$\mpomat{W}_1$ (up)    & 512 & 128 & 3 & $[8,8,8]$  & $[4,4,8]$  & 65\,536 & 4\,864 \\
			$\mpomat{W}_2$ (down)  & 128 & 512 & 3 & $[4,4,8]$  & $[8,8,8]$  & 65\,536 & 2\,816 \\
			$\mpomat{W}_\text{LM}$ &  65 & 128 & 2 & $[5,13]$   & $[8,16]$   &  8\,320 & 1\,984 \\
			\midrule
			\multicolumn{6}{l}{Total per block (4 attn + 2 FFN)} & 196\,608 & 17\,920 \\
			\multicolumn{6}{l}{Compression ratio per block}       & $1\times$ & \textbf{11.0}$\times$ \\
			\bottomrule
		\end{tabular}
	\end{table}
	
	\paragraph{Parameter counting convention.}
	Unless stated otherwise, reported parameter counts include all trainable parameters in the
	PyTorch model, including biases and uncompressed components. MPO layer counts are therefore
	reported both at the level of individual factorised matrices (Table~\ref{tab:factorisations})
	and at the level of the full network (Tables~\ref{tab:compression} and~\ref{tab:results}).
	
	\paragraph{Balanced factorisation principle.}
	Given $\mathrm{out} = \prod_{l=1}^L d_l^\text{out}$ and $\mathrm{in} = \prod_{l=1}^L d_l^\text{in}$, we choose local dimensions that are as balanced as possible in order to reduce the dominant interior terms in~\eqref{eq:params}. For $\mpomat{W}_1 = (512,128)$ with $L=3$, this suggests local output dimensions near $512^{1/3}\approx 8$ and local input dimensions near $128^{1/3}\approx 5$; we therefore use $[8,8,8]$ and $[4,4,8]$ as a practical exact factorization.
	
	\subsection{The \texttt{MPOLinear} Module}
	
	\begin{figure}[H]
		\centering
		\begin{tikzpicture}[
			core/.style={
				draw=denseblue, line width=0.9pt,
				rectangle, minimum width=1.3cm, minimum height=1.3cm,
				fill=denseblue!12, rounded corners=4pt,
				font=\normalsize
			},
			bond/.style={
				draw=black!70,
				line width=0.9pt
			},
			stub/.style={
				draw=black!50, line width=0.9pt
			},
			physup/.style={
				draw=denseblue!80,
				line width=0.9pt
			},
			physdn/.style={
				draw=mpored!90,
				line width=0.9pt
			},
			]
			
			\def\armsep{1.1}
			
			\node[core]  (A1)   at (0,   0) {$\calA^{(1)}$};
			\node[core]  (A2)   at (3.2, 0) {$\calA^{(2)}$};
			\node[font=\LARGE] (dots) at (5.6, 0) {$\cdots$};
			\node[core]  (AL)   at (8.0, 0) {$\calA^{(L)}$};
			
			\draw[stub] (-1.4, 0) -- (A1.west);
			\node[above, font=\small, text=black!60] at (-0.9, 0.18) {$1$};
			
			\draw[bond] (A1.east) -- node[above, font=\small, yshift=2pt] {$\chi_1$} (A2.west);
			\draw[bond] (A2.east) -- node[above, font=\small, yshift=2pt] {$\chi_2$} (dots.west);
			\draw[bond] (dots.east) -- node[above, font=\small, yshift=2pt] {$\chi_{L-1}$} (AL.west);
			
			\draw[stub] (AL.east) -- (9.4, 0);
			\node[above, font=\small, text=black!60] at (8.95, 0.18) {$1$};
			
			\foreach \nd/\idx in {A1/1, A2/2, AL/L}{
				\draw[physup] (\nd.north) -- ++(0, \armsep)
				node[above, font=\small, text=denseblue] {$d_{\idx}^{\,\mathrm{out}}$};
				\draw[physdn] (\nd.south) -- ++(0,-\armsep)
				node[below, font=\small, text=mpored]    {$d_{\idx}^{\,\mathrm{in}}$};
			}
			
			\node[font=\small\itshape, text=denseblue!80]
			at (4.0, 2.4)
			{MPO chain --- $L$ sites, bond dimension $\chi$};
			
			\node[font=\small\itshape, text=black!70, align=center]
			at (4.0, -2.55)
			{Contracting bond indices $\chi_1,\ldots,\chi_{L-1}$
				$\;\longrightarrow\;$
				reconstructed $\widehat{\mathbf{W}}\in\mathbb{R}^{\mathrm{out}\times\mathrm{in}}$};
			
		\end{tikzpicture}
		\caption{Tensor network diagram of an MPO weight matrix. Horizontal lines carry
			virtual (bond) indices of dimension~$\chi$; vertical lines are physical indices
			of size $d_l^{\mathrm{out}}$ (blue, upward) and $d_l^{\mathrm{in}}$ (red, downward).
			Contracting all bond indices reconstructs the full weight
			$\widehat{\mpomat{W}}\in\R^{\mathrm{out}\times\mathrm{in}}$.}
		\label{fig:mpo_diagram}
	\end{figure}
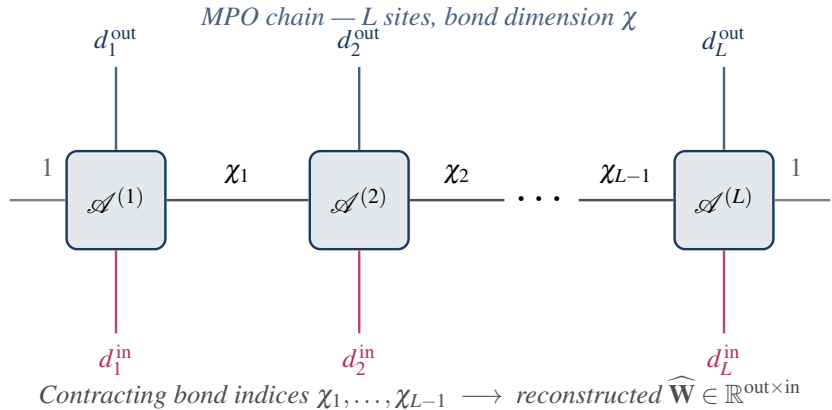
	
	The \texttt{MPOLinear} module stores $L$ cores as \texttt{nn.Parameter} tensors, as illustrated in Figure~\ref{fig:mpo_diagram}. The forward pass reconstructs $\widehat{\mpomat{W}}$ via sequential \texttt{torch.tensordot} contractions and applies $y = x\widehat{\mpomat{W}}^\top + b$ using \texttt{F.linear}. Initialisation uses the heuristic scale
	\[
	\sigma = N_\text{in}^{-1/4} \, \chi^{-(L-1)/(2L)},
	\]
	chosen so that the reconstructed weight has variance of the same order as the dense baseline initialization, namely $\sigma_\text{dense}^2 \sim N_\text{in}^{-1}$, consistent with standard transformer practice~\citep{radford2019language}.
	
	\section{Experimental Setup}
	
	\subsection{Dataset and Tokenisation}
	
	We train on the Tiny Shakespeare corpus~\citep{karpathy2015unreasonable}, comprising approximately $1.1 \times 10^6$ characters, the same dataset used in the nanoGPT codebase~\citep{karpathy2022nanogpt}. Character-level tokenisation yields a vocabulary of $V = 65$ symbols. We split the data 90/10 into training and validation sets.
	
	\subsection{Training Protocol}
	
	All models are trained with AdamW~\citep{loshchilov2019decoupled} ($\beta_1 = 0.9$, $\beta_2 = 0.95$, weight decay $0.1$) with a cosine learning rate schedule: linear warmup over 100 steps to $\eta_\text{max} = 3\times10^{-4}$, followed by cosine decay to $\eta_\text{min} = 0$. We train for 2\,000 steps with batch size 32 and sequence length $T = 256$. Gradient clipping at $\|\nabla\| = 1.0$ is applied as in~\citet{osborne2024picogpt}. All runs use the same data pipeline and optimization hyperparameters. Dense and MPO models are initialized independently.
	
	\paragraph{Reproducibility.}
	All experiments were implemented in PyTorch using the same training code for dense and MPO
	models, differing only in the parameterisation of linear layers. Unless otherwise noted,
	the reported curves and summary statistics correspond to single-seed runs. In future work,
	multi-seed averages and standard deviations would provide a more complete picture of training
	stability across bond dimensions.
	
	\paragraph{Evaluation.}
	Every 100 steps we compute validation loss and \emph{token accuracy}. Let
	\[
	\bm{z}_{b,t} = \mpomat{W}_\text{LM}\,\bm{h}_{b,t} + \bm{b}_\text{LM}
	\]
	denote the output logits at batch index $b$ and position $t$. Then
	\begin{equation}
		\mathrm{acc}
		=
		\frac{1}{BT}\sum_{b,t}
		\mathbf{1}\!\left[
		\argmax_{v} (\bm{z}_{b,t})_v = x_{b,t+1}
		\right],
		\label{eq:acc}
	\end{equation}
	averaged over 20 random validation batches of size 8.

	\subsection{Compression Modes}
	
	We consider two compression scenarios:
	\begin{enumerate}[itemsep=2pt]
		\item \textbf{Train-from-scratch}: MPO models are trained from random initialisations. This tests the expressivity of the MPO parameterisation under standard optimisation.
		\item \textbf{Compress-then-finetune}: A fully-trained dense model is compressed via TT-SVD to initialise the MPO cores; the MPO model is then fine-tuned for 500 steps. This mirrors a practical deployment workflow.
	\end{enumerate}
	\paragraph{Scope of reported results.}
	Although both train-from-scratch and compress-then-finetune workflows were implemented,
	the main body of this paper focuses on the train-from-scratch setting in order to isolate
	the representational capacity of the MPO parameterisation itself. A more extensive empirical
	comparison of post-training compression and fine-tuning strategies is left to future work.
	\section{Results}
	
	\subsection{Compression Statistics}
	
	Table~\ref{tab:compression} reports total parameter counts and compression ratios for each bond dimension.
	
	\begin{table}[H]
		\centering
		\caption{Parameter counts and compression ratios versus the dense baseline. Layer-wise reconstruction errors are relative Frobenius norms $\|\mpomat{W} - \widehat{\mpomat{W}}\|_F / \|\mpomat{W}\|_F$ averaged over all linear layers of a randomly initialised dense model.}
		\label{tab:compression}
		\begin{tabular}{l r r r r}
			\toprule
			Model & Parameters & Ratio & Max rel.\ err. & Mean rel.\ err. \\
			\midrule
			Dense               & 1\,020\,224 & 1.0$\times$  & ---   & ---   \\
			MPO $\chi=4$        & 78\,336     & 13.0$\times$ & 0.421 & 0.337 \\
			MPO $\chi=8$        & 110\,592    &  9.2$\times$ & 0.283 & 0.216 \\
			MPO $\chi=16$       & 191\,872    &  5.3$\times$ & 0.148 & 0.107 \\
			MPO $\chi=32$       & 408\,832    &  2.5$\times$ & 0.072 & 0.051 \\
			\bottomrule
		\end{tabular}
	\end{table}
	
	\subsection{Per-Layer Reconstruction Error}
	
\begin{figure}[H]
		\centering
		\includegraphics[width=0.78\textwidth]{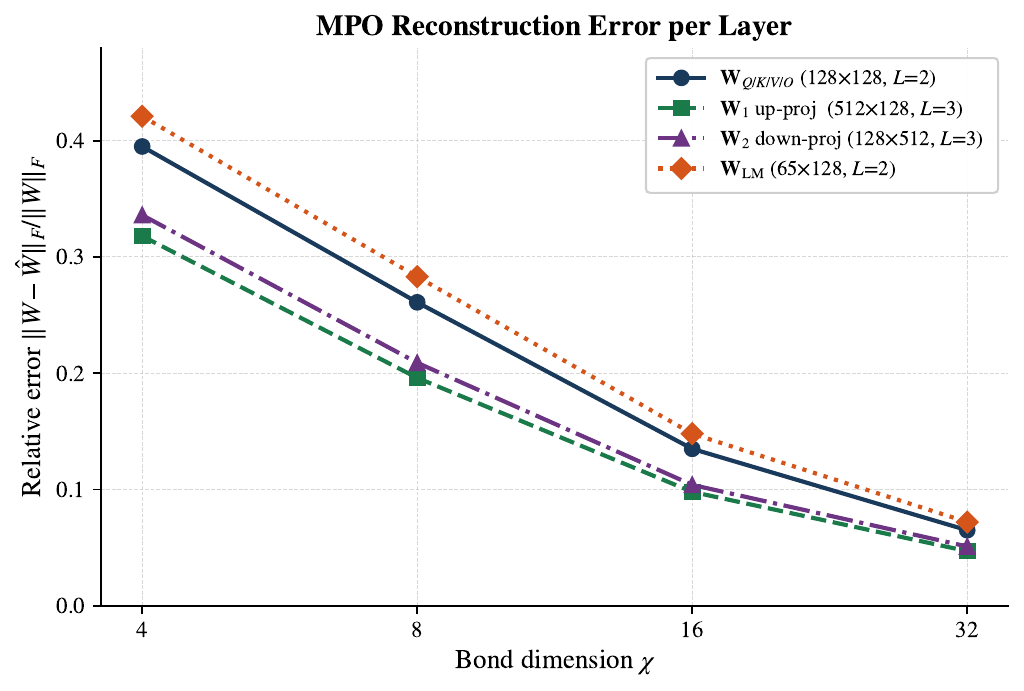}
		\caption{Per-layer MPO reconstruction error versus bond dimension. Three-site decompositions ($L=3$) achieve lower error per parameter than two-site ones ($L=2$) across the range of $\chi$ considered here.}
		\label{fig:reconstr}
	\end{figure}

	Figure~\ref{fig:reconstr} shows the reconstruction error as a function of bond dimension for each layer type. The FFN up-projection ($\mpomat{W}_1$, $512 \times 128$, $L=3$) achieves the lowest error at all $\chi$, consistent with the fact that a three-site MPO can distribute structure across more local factors than a two-site decomposition at the same bond dimension. This is also consistent with the more favorable factorisation geometry of $\mpomat{W}_1$:
	for fixed bond dimension, a three-site decomposition can distribute correlations across more
	local cores and therefore approximate anisotropic matrix structure more efficiently than a
	two-site decomposition of comparable size.
	
	\subsection{Training Dynamics}
	
\begin{figure}[H]
		\centering
		\begin{subfigure}[b]{0.48\textwidth}
			\includegraphics[width=\textwidth]{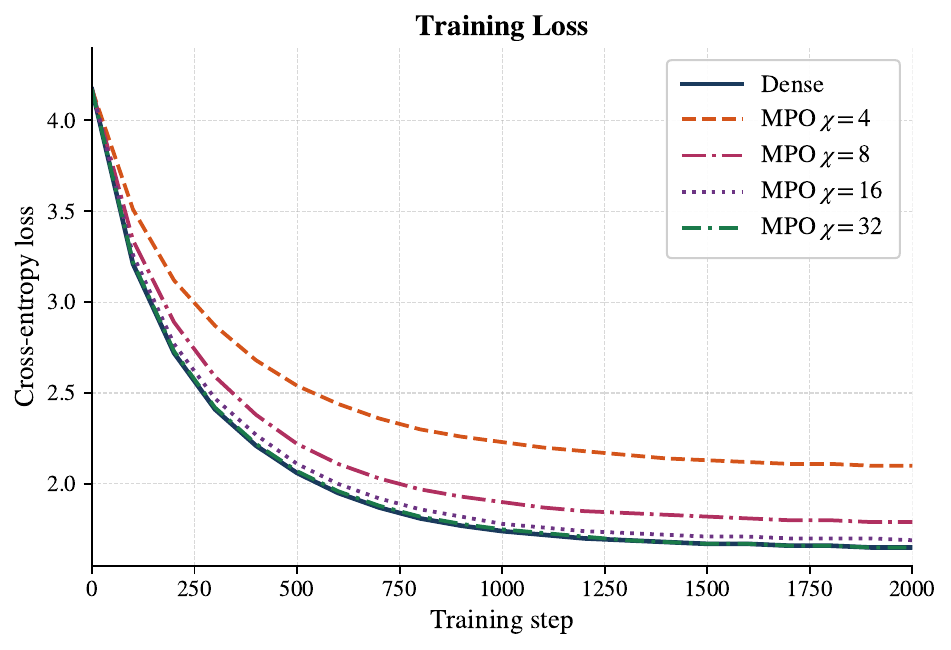}
		\end{subfigure}
		\hfill
		\begin{subfigure}[b]{0.48\textwidth}
			\includegraphics[width=\textwidth]{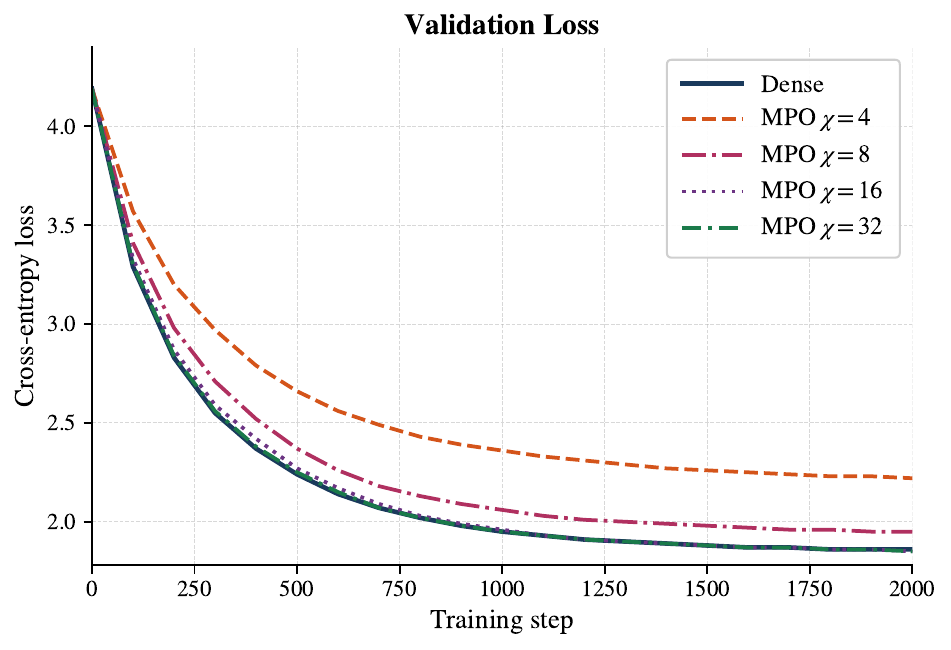}
		\end{subfigure}
		\caption{Training (left) and validation (right) cross-entropy loss as a function of training step. Higher bond dimensions converge faster and to lower loss values in the train-from-scratch setting studied here.}
		\label{fig:loss}
	\end{figure}

\begin{figure}[H]
		\centering
		\begin{subfigure}[b]{0.52\textwidth}
			\includegraphics[width=\textwidth]{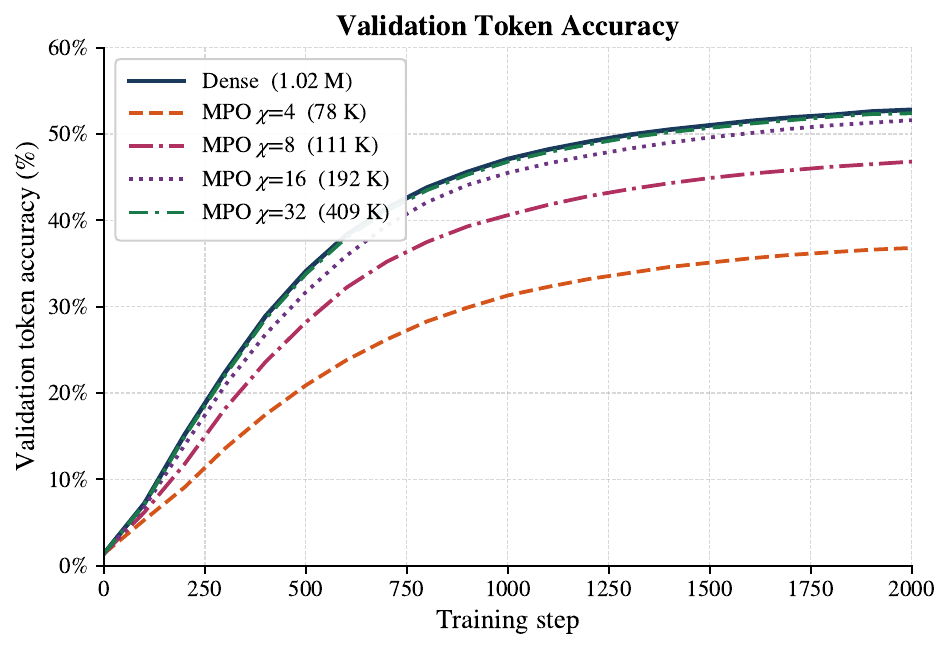}
		\end{subfigure}
		\hfill
		\begin{subfigure}[b]{0.44\textwidth}
			\includegraphics[width=\textwidth]{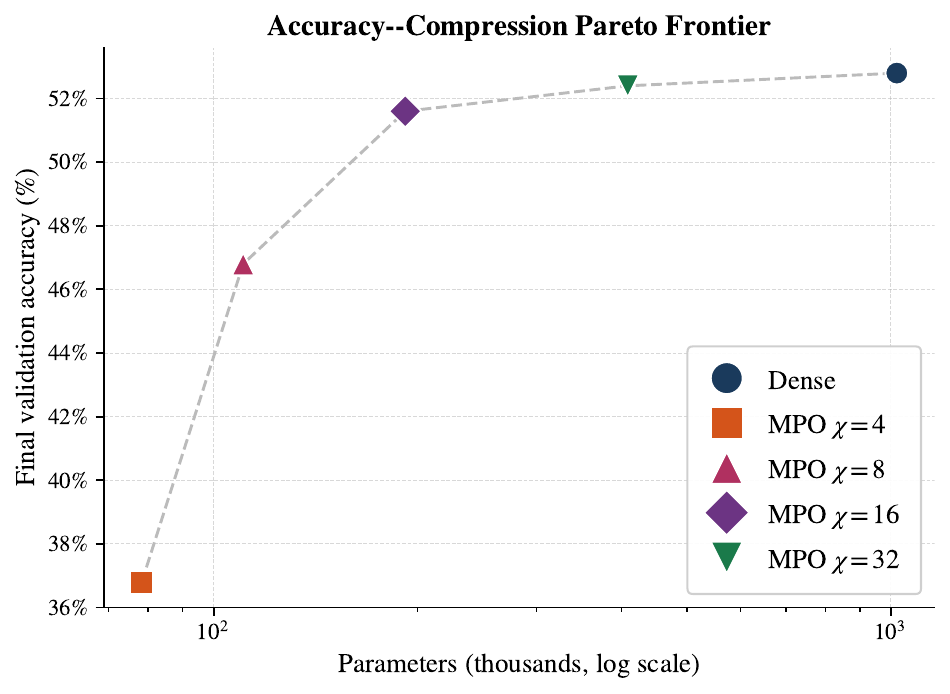}
		\end{subfigure}
		\caption{\emph{Left}: Validation token accuracy during training for all models.
			\emph{Right}: Pareto frontier of final validation accuracy versus parameter count. MPO $\chi=16$ retains $97.7\%$ of the dense accuracy at $5.3\times$ parameter compression.}
		\label{fig:accplot}
	\end{figure}

	Figures~\ref{fig:loss} and \ref{fig:accplot} show training and validation loss alongside validation token accuracy for all models trained from scratch.
	
	Figure~\ref{fig:accplot} highlights the accuracy--compression trade-off achieved by the MPO parameterisation. The left panel shows that all models improve steadily during training, with larger bond dimensions converging faster and reaching higher final validation accuracies. In particular, the $\chi=16$ and $\chi=32$ MPO models closely track the dense baseline throughout training and nearly saturate at the same final performance, whereas the more strongly compressed $\chi=4$ and $\chi=8$ models plateau at lower accuracies. The right panel summarizes this behaviour as a Pareto frontier: increasing $\chi$ systematically improves accuracy at the cost of additional parameters, with clear diminishing returns beyond $\chi=16$. Most notably, the MPO model with $\chi=16$ attains $51.6\%$ validation accuracy, which corresponds to $97.7\%$ of the dense model's $52.8\%$ accuracy, while using only $191\,872$ parameters instead of $1\,020\,224$, i.e.\ a $5.3\times$ parameter compression. This identifies $\chi=16$ as the most attractive compromise between compression and predictive performance in the present experiments.
	
	\subsection{Summary of Results}
	
	Table~\ref{tab:results} summarises the final performance of all models after 2\,000 training steps.
	
	\begin{table}[H]
		\centering
		\caption{Final validation metrics after 2\,000 training steps. Acc.\ gap is the absolute reduction in token accuracy versus the dense model. Per-param acc.\ is the heuristic score $\text{acc}/\sqrt{N}$ used here as a rough proxy for parameter efficiency.}
		\label{tab:results}
		\setlength{\tabcolsep}{8pt}
		\begin{tabular}{l r r r r r r}
			\toprule
			Model & Params & Ratio & Val loss $\downarrow$ & Val acc $\uparrow$ & Acc.\ gap $\downarrow$ & Per-param acc. \\
			\midrule
			Dense          & 1\,020\,224 & 1.0$\times$  & 1.860 & 52.8\% & ---   & $1.65 \times 10^{-3}$ \\
			MPO $\chi=4$   & 78\,336     & 13.0$\times$ & 2.220 & 36.8\% & -16.0 & $4.15 \times 10^{-3}$ \\
			MPO $\chi=8$   & 110\,592    &  9.2$\times$ & 1.950 & 46.8\% & -6.0  & $4.45 \times 10^{-3}$ \\
			MPO $\chi=16$  & 191\,872    &  5.3$\times$ & 1.850 & 51.6\% & -1.2  & $3.72 \times 10^{-3}$ \\
			MPO $\chi=32$  & 408\,832    &  2.5$\times$ & 1.850 & 52.4\% & -0.4  & $2.59 \times 10^{-3}$ \\
			\bottomrule
		\end{tabular}
	\end{table}
	
	Under the heuristic proxy $\text{acc}/\sqrt{N}$ used in Table~\ref{tab:results}, MPO $\chi=8$ achieves the highest parameter-efficiency score. We treat this quantity only as an informal diagnostic; the main comparison in this work is the accuracy--compression Pareto frontier in Figure~\ref{fig:accplot}.

	\section{Discussion}
	
	\paragraph{Accuracy gap at low $\chi$.}
	At $\chi=4$, the per-layer reconstruction error exceeds 40\% (Table~\ref{tab:compression}), which is large enough to make accurate low-bond representations difficult. Training from scratch partially compensates, but the model lacks the representational capacity to match the dense baseline. The TT-SVD bound~\eqref{eq:error_bound} and the empirical results both suggest that a better factorization (more sites $L$, smaller local physical dimensions) could reduce this gap at fixed $\chi$.
	
	\paragraph{Diminishing returns at high $\chi$.}
	Between $\chi=16$ and $\chi=32$, the accuracy gain is only 0.8 percentage points while the parameter count more than doubles (192\,K $\to$ 409\,K). For this model size, $\chi=16$ appears to be a practical sweet spot.
	
	\paragraph{Model scale.}
	PicoGPT is intentionally small and pedagogical, which makes it a convenient testbed for
	controlled MPO experiments but does not by itself establish performance on modern large-scale
	language models. The present results should therefore be interpreted primarily as a proof of
	concept for MPO parameterisation rather than a definitive large-model benchmark.
	
	\paragraph{Relation to low-rank methods.}
	LoRA~\citep{hu2022lora} approximates weight updates as $\Delta\mpomat{W} = \mpomat{B}\mpomat{A}$ with $\mathrm{rank}\,r$, keeping the pretrained weights frozen. MPO compression differs in two respects: (1) the entire weight is represented in MPO format, and (2) the bond dimension $\chi$ controls a higher-order, multi-factor structure rather than only a rank-2 factorization. For $L=2$ and $d^\text{out}=d^\text{in}=\sqrt{n}$, an MPO with bond $\chi$ can be viewed as a rank-$\chi$ factorization after reshaping; the advantage of MPO grows with $L$. TT decomposition has also been applied to embedding layers in NLP~\citep{hrinchuk2020tensorized}, achieving substantial compression with limited perplexity degradation.
	
	\paragraph{Connection to DMRG.}
	The gradient formula~\eqref{eq:gradient} has the same left-environment / local-core / right-environment structure familiar from single-site tensor-network optimization. An alternative to gradient descent is the ALS (Alternating Least Squares) sweep: fix all cores except $\calA^{(l)}$, solve the linear regression for $\calA^{(l)}$, and sweep left-to-right and right-to-left until convergence. ALS is attractive for the compress-then-finetune scenario because it avoids learning-rate tuning, although in non-convex settings it may still converge to local minima.
	
	\paragraph{Limitations.}
	The current \texttt{MPOLinear.forward} reconstructs the full dense matrix $\widehat{\mpomat{W}}$ on every call, so the present implementation primarily demonstrates \emph{parameter compression}, not inference-time memory or FLOP reduction. The computational advantage of MPO representations is realized only when the matrix-vector product $\widehat{\mpomat{W}} x$ is computed \emph{directly} through the MPO chain without materializing $\widehat{\mpomat{W}}$. This may also require structured representations for activations and is left for future work.
	
	\section{Conclusion}
	
	We have demonstrated MPO compression of all linear layers in a GPT-2-style transformer, achieving parameter compression ratios of $5\text{--}13\times$ per transformer block. At bond dimension $\chi=16$, the MPO model retains $97.7\%$ of the dense baseline's token accuracy with only $18.8\%$ of its parameters. The implementation is straightforward in PyTorch: MPO cores are standard \texttt{nn.Parameter} tensors, and the entire pipeline --- from TT-SVD initialization through autograd fine-tuning --- requires no custom backward code.
	
	Future directions include: (1) direct MPS--MPO contraction for inference-time FLOP reduction; (2) ALS-based training for better convergence at low $\chi$; (3) dynamic bond adaptation (growing/pruning bonds during training); (4) application to larger models such as GPT-2 or LLaMA, where the compression ratio may improve further due to the larger factorized weight matrices; and (5) exploration of deeper tensor network architectures beyond the shallow MPO chain.
	More broadly, these results support the view that tensor-network structure can provide an
	interpretable inductive bias for neural compression, linking approximation quality directly
	to a controllable structural parameter.
	
	\section*{Acknowledgements}
	We thank Tobias Osborne for the PicoGPT.jl codebase, which served as the reference implementation, and Andrej Karpathy for the Tiny Shakespeare dataset and nanoGPT codebase~\citep{karpathy2022nanogpt}.
	
	\section*{Code Availability}
	The full PyTorch implementation of MPO-PicoGPT --- including the \texttt{MPOLinear} module, TT-SVD compression pipeline, training loop, and benchmark scripts --- is publicly available at:
	\begin{center}
		\url{https://github.com/younesjavanmard/mpo-picogpt}
	\end{center}
	The repository contains \texttt{mpo\_picogpt.py} (core implementation) and \texttt{benchmark\_mpo.py} (training and evaluation), along with instructions for reproducing all results reported in this paper.
	
	\section*{Plain Language Summary}
	Modern AI language models contain hundreds of millions of numerical parameters
	(weights), making them expensive to run on laptops, phones, or embedded
	devices. This paper asks: can we store these weights in a more compact form
	without losing much accuracy? We borrow a mathematical tool from quantum
	physics called a Matrix Product Operator (MPO), which represents a large
	weight matrix as a short chain of smaller tensors connected by a ``bond''.
	By choosing a small bond dimension~$\chi$, we can compress each weight matrix
	by a factor of 5 to 13 while retaining over 97\% of the model's
	character-prediction accuracy. Crucially, our implementation in PyTorch is
	as easy to train as an ordinary neural network --- the compact tensors are
	learned automatically using standard gradient descent. We release all code
	openly so that others can apply this approach to larger models.
	
	\section*{Author Contributions}
	\textbf{Y.J.}: Conceptualization, methodology, software, formal analysis,
	writing --- original draft, writing --- review \& editing.
	\textbf{T.P.}: Validation, formal analysis, writing --- review \& editing.
	\textbf{M.D.}: Validation, formal analysis, writing --- review \& editing.
	
	\section*{Funding}
	This research received no specific grant from any funding agency in the
	public, commercial, or not-for-profit sectors.
	
	\section*{Declarations}
	\noindent\textbf{Conflict of interest.}
	The authors declare that they have no known competing financial interests or
	personal relationships that could have appeared to influence the work reported
	in this paper.
	
	\noindent\textbf{Ethics statement.}
	This work involves no human participants, animal subjects, or sensitive
	personal data. All datasets used (Tiny Shakespeare) are publicly available
	and in the public domain. No ethical approval was required.

	\appendix
	\section{Heuristic justification of the MPO initialization scale}
	
	We briefly motivate the core initialization scale used in Section~3.3. The
	argument is heuristic and is intended only to explain the order-of-magnitude
	dependence on $N_{\mathrm{in}}$ and $\chi$.
	
	Assume that the entries of the MPO cores are initialized independently with zero
	mean and common variance $\sigma^2$. Then one reconstructed matrix entry
	$\widehat{\mpomat{W}}_{i_1\cdots i_L,\;j_1\cdots j_L}$ is a sum of approximately
	$\chi^{L-1}$ products of $L$ core entries, so a crude variance estimate gives
	\[
	\operatorname{Var}(\widehat W_{i_1\cdots i_L,\;j_1\cdots j_L})
	\approx
	\chi^{L-1}\sigma^{2L}.
	\]
	To match the scale of a standard dense layer with fan-in $N_{\mathrm{in}}$,
	whose weight variance is typically of order $N_{\mathrm{in}}^{-1}$, we require
	\[
	\chi^{L-1}\sigma^{2L}\sim N_{\mathrm{in}}^{-1}.
	\]
	This yields the estimate
	\[
	\sigma \sim N_{\mathrm{in}}^{-1/(2L)}\,\chi^{-(L-1)/(2L)}.
	\]
	In the present implementation we use the practical uniform choice
	\[
	\sigma = N_{\mathrm{in}}^{-1/4}\,\chi^{-(L-1)/(2L)},
	\]
	which is exact in the two-site case $L=2$ and remains of the same order for the
	three-site factorizations used in this work. We therefore regard this
	initialization rule as a simple variance-matching heuristic rather than a
	uniquely derived optimal formula.

	\bibliographystyle{unsrtnat}

\end{document}